\documentclass[10pt,journal,compsoc]{IEEEtran}
\usepackage{enumerate}
\usepackage{mathtools}    

\usepackage{algorithm}
\usepackage{algorithmic}
\usepackage{color,soulutf8}
\usepackage{xcolor}

\usepackage{amsmath} 
\usepackage{amsthm}
\usepackage{bm}
\usepackage{graphicx}
\usepackage{amsfonts}
\usepackage{relsize}
\usepackage{float}
\usepackage{subfigure}
\usepackage{subfigmat}
\usepackage{etoolbox}
\usepackage{booktabs}

\patchcmd{\subfigmatrix}{\hfill}{\hspace{0.8cm}}{}{}

\usepackage{amsmath,amsfonts,amssymb}

\begin{document}

%

\title{Spatiotemporal Filtering for Event-Based Action Recognition}



%
\author{\IEEEauthorblockN{
Rohan Ghosh\IEEEauthorrefmark{1}
Anupam Gupta\IEEEauthorrefmark{2}
Andrei Nakagawa\IEEEauthorrefmark{2}
Alcimar B. Soares \IEEEauthorrefmark{2}
Nitish V. Thakor\IEEEauthorrefmark{2}}
\IEEEauthorblockA{\IEEEauthorrefmark{1}School of Electrical and Computer Engineering\\
National University of Singapore,
Singapore}
\IEEEauthorblockA{\IEEEauthorrefmark{2}Singapore Institute for Neurotechnology\\
National University of Singapore,
Singapore}

}




\IEEEtitleabstractindextext{%
\begin{abstract}
In this paper, we address the challenging problem of action recognition, using event-based cameras. To recognise most gestural actions, often higher temporal precision is required for sampling visual information. Actions are defined by motion, and therefore, when using event-based cameras it is often unnecessary to re-sample the entire scene. Neuromorphic, event-based cameras have presented an alternative to visual information acquisition by asynchronously time-encoding pixel intensity changes, through temporally precise spikes ($\approx 10 \ \mu$s resolution), making them well equipped for action recognition. However, other challenges exist, which are intrinsic to event-based imagers, such as higher signal-to-noise ratio, and a spatiotemporally sparse information. One option is to convert event-data into frames, but this could result in significant temporal precision loss. In this work we introduce spatiotemporal filtering in the spike-event domain, as an alternative way of channeling spatiotemporal information through to a convolutional neural network. The filters are local spatiotemporal weight matrices, learned from the spike-event data, in an unsupervised manner. We find that appropriate spatiotemporal filtering significantly improves CNN performance beyond state-of-the-art on the event-based DVS Gesture dataset. On our newly recorded action recognition dataset, our method shows significant improvement when compared with other, standard ways of generating the spatiotemporal filters.
\end{abstract}

\begin{IEEEkeywords}
Event-based vision, Feature extraction, Slow feature analysis, Feature point tracking
\end{IEEEkeywords}}

\maketitle


%

\IEEEraisesectionheading{\section{Introduction}\label{sec:introduction}}

Neuromorphic visual sensors \cite{dvs_original,atis_original,davis_original} present an event-based, asynchronous approach to visual information acquisition. This is in contrast to frame-based cameras which synchronously sample entire image arrays. Event based sensors only respond to changes in the visual scene, by generating precisely timed \textit{spike-events} at the corresponding pixels where the pixel intensity changed by a threshold amount. Additionally, due to monitoring \textit{relative} pixel intensity changes, sensor response is stable across a wide range of illumination in the scene, showing a dynamic range of operation. Each asynchronous pixel outputs information only when a threshold amount of relative pixel intensity change is noted. This helps avoid considerable redundancy in visual information. The times of changes are recorded in a very high temporal precision (10 $\mu$s), enabling high-speed operation. Note that such temporal precision of 10 $\mu$s can only be achieved in frame-based cameras capturing 100,000 frames in a second, which requires massive data storage and computational capabilities to process. Therefore event-based sensors achieve the impressive task of being high-speed and low throughput at the same time \cite{dvs_original}. 


In this paper we primarily tackle gesture and action recognition problems, with event-based cameras. For a system to categorize actions or gestures, it primarily needs both static scene information and dynamic motion information. For a static event-based camera facing a static background, dynamic change information should be sufficient for identifying human actions. Moreover, filtering out all background information and only keeping foreground motion is of obvious benefit for this problem. Such properties, coupled with high temporal precision in recording motion-related scene changes, makes event-based cameras well suited for such applications. Here we tackle the problem of spatiotemporal pattern recognition with event-data obtained from such cameras. 

 Before addressing spatiotemporal classification, we look into the important characteristics of classifiers handling static, frame-based images for recognition problems. Over the last decade, convolutional neural networks have proved themselves to be very astute for this task when presented with enough data. One can broadly attribute the success of 2D convolutional neural networks (CNN,  \cite{cnn_digit}) to the following reasons. 
\begin{itemize}
\item \textit{Hierarchical Representation}: Deep CNNs  extract low-level edge like features in the lower layers, and high level entity presence/absence features in the higher layers ( \cite{cnn_layerwise}). Having the capability to form hierarchical representations is crucial for a classifier ( \cite{hierarchy_recognition} paper that proposes compositionality/hierarchy as important classification properties).
\item \textit{Easier to achieve translation invariance}: Since each convolution operation is usually repeated throughout the whole image, convolutional filters look for a certain feature in the whole image array, independent of the feature. The max-pooling layers allow the CNN to only channel the maxima of a convolution output, which is invariant of the location of the corresponding feature within the image. This doesn't imply that the CNN is translation invariant by default, as often data augmentation methods like translation are required to enhance performance and prevent overfitting (including in this paper). 
\item \textit{High-level features are global, low-level features are local}: Even deep neural networks (DNNs), which are fully connected, have a hierarchical representation, as higher layer features are naturally represent more complex abstractions of the input. CNNs on the other hand can additionally boast of such hierarchical representations being organized such that the low-level representations detect small-scale local features (edges, corners) whereas the high-level, deeper responses of the CNN detect coarse features spread out over a larger image region (shapes, objects, parts of objects) \cite{cnn_layerwise}. 
\end{itemize}

Inspired by the success of the 2D CNN learning framework, similar 3D convolutional neural network architectures have been proposed for spatiotemporal classification problems. This includes problems such as human action recognition ( \cite{cnn3d_action}),  identifying sports ( \cite{cnn_sports1m}) and gesture recognition ( \cite{cnn3d_gesture} . Unlike the 2D case, 3D CNNs deal with spatiotemporal data by replacing 2D convolutions with 3D space-time convolution kernels. Different from a 2D convolution, which usually extracts edge orientation, edge gradient etc. ( \cite{cnn_transferable_bengio}), a 3D spatiotemporal convolution will extract features incorporating both edge and motion information. Thus 3D convolutional networks retain the aforementioned properties of a 2D CNN.

To make CNNs work with event-based data, one first needs to structurize the spike-events, which arrives asynchronously as an ever-growing set of three dimensional points. To convert the spike-event point cloud data obtained from a neuromorphic camera into a structured matrix form, one approach is to simply create time-surface images ( \cite{cnn_ts_steering,hots_garrick}), or more commonly, binned event-count images (\cite{mapping_frame_event,cnn_rohan}). For recognizing spatiotemporal patterns, the input usually consists of multiple consecutive images, that serve as channels, (or, less effectively, separately  \cite{cnn_action_separate}) to generate a reliable class estimate. In practice, a spatiotemporal pattern like a hand gesture needs $\sim 250$ ms of visual information to be reliably recognized. 

\begin{figure*}[htbp]
	\centering
	\includegraphics[width=0.8\textwidth]{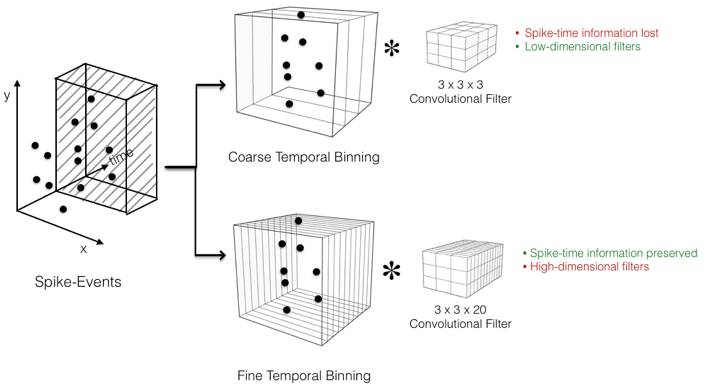}
	\caption{{Two} possible approaches to adapting 3d convolutions for event-based data input, along with the tradeoffs (green and red) for each approach. More discussed in Section \ref{sec:introduction}}
	\label{fig:coarse_fine_cnn}
\end{figure*}

Yet another possible approach to structurizing the spike-event data is to create a 3D grid of voxels, which form a 3D spatiotemporal matrix, where each voxel contains the number of spike-events within it (see Fig. \ref{fig:coarse_fine_cnn}). In this case, 3D convolution based filters can be seamlessly integrated to work with such spatiotemporal data. Since we wish to preserve temporal acuity of the spike-events, smaller voxels with smaller time dimension have to be chosen. In doing so however, the size of the convolution kernels also increase. Typically, $(3\times 3 \times 3)$ spatiotemporal convolution kernels have been found to work well with frame-based data ( \cite{cnn_3d_simple}). To reproduce such a kernel for spike-event data, we need a convolutional kernel which typically spans $30 \times 3 = 90$ ms, with three partitions along the temporal dimension. This leads to loss of temporal precision of each spike-event up to $\pm 15$ms, which is a considerable loss of spike-timing precision, given the original precision can be as low as tens of microseconds. To reduce this information loss, a possible workaround is to increase the number of time partitions by decreasing the voxel size along the temporal dimension. For example, if each partition was about 2 ms, then the size of the convolutional filters would be $3 \times 3 \times (90 \ ms / 2 \ ms) = (3 \times 3 \times 45)$. This points to a 15 fold increase in the number of parameters per filter, which is really inefficient for a learner, and might lead to overfitting. {(\cite{cnn_weight_dimensionality})} 

In this work, we wish to adapt the 3D CNN architecture to function on asynchronous spike-event data streams, by addressing the above concerns. Our proposed architecture divides the system into two-stages. Our main contribution is the unsupervised algorithm for learning the spatiotemporal filters present in the initial pre-processing stage. The filter learning algorithm maximizes the robustness of the spatiotemporal filters to spike-event pattern distortions, while preserving as much of the information present within the local spatiotemporal regions\footnote{Which are of the same size as the convolutional filters}. For e.g., consider a certain visual pattern which is responsible for a set of spike-events within a local spatiotemporal area. The key observation is that predictable modifications on the feature pattern involving: \textit{edge sharpness, motion magnitude, illumination level} must elicit minimal changes to the feature values.

The second, supervised stage, consists of a convolutional neural network to which the spatiotemporal filter responses are input as feature channels. The spatiotemporal filters are expected to channel useful information to the supervised classifier to enhance classification performance. We test the performance of the system on the gesture recognition dataset ( \cite{gesture_ibm}) and an action recognition dataset that we recorded, which contains six actions (similar to  \cite{action_data_kth}). On the gesture recognition dataset, system is able to achieve 93.5 \% gesture recognition accuracy (11 classes) with only 50 ms of event data.  With larger input duration (100 ms) we achieve 94.8\% recognition accuracy. Note that a frame-based camera would need at least 2 frames (66 ms duration for a 30 FPS camera) to get motion information. Since event-based cameras output visual change information continuously, often 50 ms of data is enough to estimate motion parameters \cite{flow_ryad}. Moreover, the average latency of the system from the onset of a gesture, in identifying the correct gesture, is only 56 ms.

\section{Related Work}

Our proposed method supplements the conventional CNN architecture with an unsupervised filtering stage at the beginning which extracts feature responses. There are other examples of a similar structure being used for neuromorphic vision  \cite{cnn_orientation_filters}. In that work, hand-crafted orientation detection filters were used to create multiple input channels, which were input to a CNN for classifying objects. The authors were able to obtain state-of-the-art performance with this methodology on two benchmark event-based vision datasets. {In \cite{hfirst}, two sequential layers of filtering, involving simple feature detectors (orientation) and then subsequently complex features (corners, curves etc.) were used for static recognition problems with the DVS.} Their system was asynchronous, as the filter responses were updated with each incoming spike-event. However, the finally obtained features were subject to template matchers for the final recognition step, as opposed to another classifier architecture as we propose here.

It remains to be seen though whether temporal information present in the spike-events can be constructively used for static recognition problems. For recognizing spatiotemporal patterns, i.e. action recognition, spike-time information is crucial. The bottleneck on recognition accuracy in part depends on the use of spike-timing information in a classifier. One of the prominent examples is in event-based gesture recognition with spiking convolutional neural networks in \cite{gesture_ibm}. A total of 10 categories of gestures were recorded in three different illumination configurations for 29 participants. This is one of the datasets on which our algorithm performance is tested in this work. The system achieved 94.5\% recognition accuracy for 11 categories, when each classifier decision was made with 192 ms of input event-data \footnote{The original system channels spike-events spread across a duration of 32 + 16*5 = 112 ms (see \cite{gesture_ibm}). However, the final accuracies were obtained after a sliding window filter on the network responses across 80 ms. This indicates that the time duration of spikes involved having a non-zero impact in the final decision that led to highest accuracy (94.5\%) was 112+80 = 192 ms.}. Optical flow based features were extracted in \cite{motion_feature_event}, which were later combined with a bayesian framework, to perform gesture recognition for simple hand gestures.

Our method is the first example in the field where the hierarchical learning structure of a CNN is combined with robust spatiotemporal filters learnt in an unsupervised manner. Our system betters state-of-the-art performance by $0.3\%$, with half as much spike-event input (100 ms). Our method learns projections which are invariant to stochastic removal of the spike-events, and is able to outperform PCA and random projections by a large margin (20\%), when considering classifier performance on our recorded action recognition dataset.

\section{Spatiotemporal Filters}

\section{Methods} \

The DVS produces spike-events $e_i = (x_i,y_i,t_i)$ at a rate of approximately 15k sec$^{-1}$. Each event also contains polarity information, but we do not include polarity information in feature creation in this work. The following sections mention the details of the two stages (unsupervised filter learning and CNN based classification) of our system.

\begin{figure*}[h]
\centering
\includegraphics[width=0.9\textwidth]{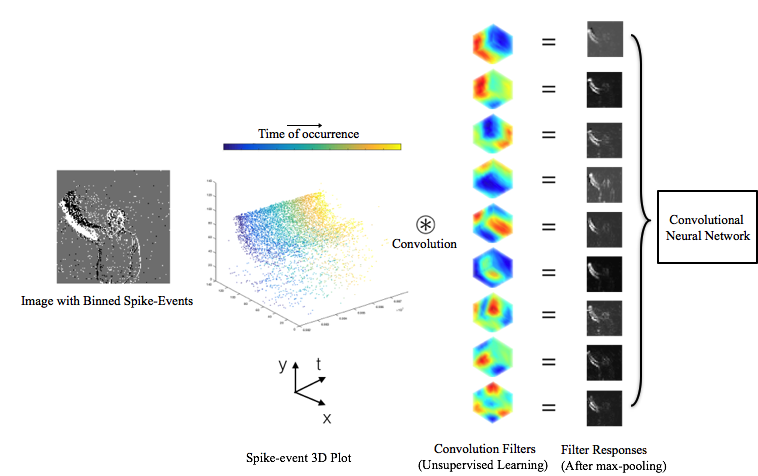}
\label{fig:2stage_overview}
\caption{The two-stage processing pipeline of the method described in this work is shown with real event-data input, of a person hand-waving at the camera. The raw event-data is convolved with 3D spatiotemporal filters to generate multiple filtered channels, as the input to a convolutional neural network classifier. The spatiotemporal filters are learnt separately, in an unsupervised manner, as detailed in Section \ref{sec:learn_filters}. Shown are the filters learnt with ROI side $6\times 6\times 50$, covering a temporal range of 100 ms. Notice how each filter responds maximally to a different part of the input.}
\end{figure*}

\subsection{Spatiotemporal Filter Learning} \label{sec:learn_filters}


The spatio-temporal space of events is first partitioned into a three dimensional grid of voxels. Each voxel is a box shaped ROI of dimension $1$ pixel $\times$ $1$ pixel $\times$ $t_{vox}$ $\mu s$. This implies that a 3D Matrix $V$, of size $128 \times 128 \times N$ is constructed, where $N = (t_N - t_1)/t_{vox}$. Each element of $V$ indicates the number of events inside the corresponding voxel location. Given a set of events $\textbf{E}$, let $V\left ( \textbf{E}, t_{vox} \right )$ denote the spike count matrix thus generated when the time dimension of each voxel is $t_{vox}$.

The 3D matrix $V$ created as a result of the voxellation is naturally suitable for convolution operations on it. To that end, we define a spatiotemporal filter $F_{ST}$ to be a 3D convolutional kernel of size $(a \times a \times k)$, where $(a \times a)$ is its spatial size and $k$ the time dimension (see Fig. \ref{fig:2stage_overview}). Since in this work we don't handcraft such filters, $F_{ST}$ is instead estimated from the event-data. To do so, from the training data we randomly sample subregions from $V\left ( \textbf{E}, t_{vox} \right )$, which are denoted by $X_1,X_2,...,X_M$, which are of the same size as $F_{ST}$. Note that the actual size of these subregions is $(a \times a \times kt_{vox})$. {We denote the temporal window length by $T=kt_{vox}$}. In our experiments, typical values of the size parameters of $F_{ST}$ are $a \in (4 \sim 10)$, $k \in (50 \sim100)$ and $t_{vox} \in (2 \sim 5) \ ms$, which sets the dimensionality of $F_{ST}$ to the range $(1000 \sim 10000)$. Due to this high dimensionality, we impose extra constraints in the filter response structure, through an unsupervised learning algorithm. The constraints imposed on the unsupervised learning of $F_{ST}$ are as follows. 

\begin{itemize}
\item \textbf{Information preservation:} Like principal components analysis (PCA), the filters must preserve the highly variant aspects of the data. Ideally we shouldn't have filters which give very similar responses for all input data points. This is the most basic dimensionality reduction tool that can be employed for such a problem. 
\item \textbf{Robustness to event removals:}  Events generated are probabilistic in nature, i.e. any particular event might not recur in another recording of the same input stimulus. Also note that event-patterns are very susceptible to illumination level around a visual feature. It is therefore desired to compute spatiotemporal filters which give similar responses when a few of the spike-events are randomly removed from a local spatiotemporal neighborhood. 
\end{itemize} 

Merging the above two constraints leads us to a SFA-like optimization which minimizes the following quantity (slowness parameter), 

\begin{equation} \label{eq:to_minimize1}
\mathbb{E}_i  \left [ \left (F_{ST}^{\intercal}X_i - F_{ST}^{\intercal}\phi(X_i) \right ) ^2 \right] ,
\end{equation}

subject to
\begin{equation}\label{eq:to_minimize2}
\mathbb{E}_i \left [ \sum_{i = 1}^{M} \left (F_{ST}^{\intercal}X_i -  \mu \left( F_{ST}^{\intercal} X_i \right) \right ) ^2 \right ] = 1
\end{equation}

The above optimization will return a set of orthogonal filters, which iteratively minimize the value of the slowness parameter. Here, $\phi(X)$ is a function that simulates the removal of an event contained in $X$. The process of obtaining $\phi(X)$ from $X$ is outlined in Algorithm 1. Note that both of the discussed constraints have been incorporated into one objective function. The numerator enforces the filter responses to be similar across event removals, and the denominator ensures that the filters encode sufficiently variance in the projections of the input. Optimization of the above function is straightforward, as detailed in the original work that proposed SFA \cite{sfa_original}. The top-n spatiotemporal filters, i.e. the filters with the n smallest values of the slowness parameter, often have quite interesting patterns reflective of their spatiotemporal structure (see Fig. \ref{fig:filter_shape_temporal}).

\begin{algorithm} 
 \caption{Computation of $\phi(X)$ from $X$}
 \label{alg:sparse_to_dense}
\begin{algorithmic} 
 \REQUIRE Given the set of events $e_i = (x_i,y_i,t_i)$,  $ 1 \leq i \leq N$, which are present in the spike-count matrix $X$.
 \ENSURE The modified spike-count matrix $\phi(X)$ 
 \STATE \textbf{for} every event $e_i$ \textbf{do} \\
 \STATE   \ \ \ \  Remove $e_i$ with its nearest neighbor $e_{NN(i)}$
 \STATE  \textbf{endfor}
 \STATE Construct the spike-event count matrix $\phi(X)$ with $\{e_{NN(i)} \} _{i=1}^{N}$
\end{algorithmic}
\end{algorithm}

\subsection{Filter response input to CNN} 
 
Once the top-n spatiotemporal filters are learned from the available event data (training), we use them as a pre-processing stage to a CNN. Let us denote the top-n filters as $\left(F^{1}_{ST},F^{2}_{ST}, \ ... \ ,F^{n}_{ST}\right )$, each of size $(a \times a \times k)$. First, the spike count matrix obtained from the entire event recording, $V\left ( \textbf{E}, t_{vox} \right )$, of size $(128 \times 128 \times N)$, is partitioned into $N/k$ matrices, each of size $(128 \times 128 \times k)$.{ This is done so that the time dimension of each partition equals the time dimension of each convolutional filter. Each spike count matrix generated from the corresponding partition, is convolved with each filter, to generate a total of $n$ filter response images, $I_1, I_2, ..., I_n$, of size $\left((128-a +1) \times (128- a +1) \times 1 \right)$. The filter output is normalized, so that the mean of the convolution output of each filter is $0$, and the variance is $1$ (when computed on all available training data). This generates a modified, normalized filter responses $\tilde{I} _1, \tilde{I} _2, ..., \tilde{I} _n$ (which are images). Each filter's output is then passed through a hyperbolic tangent function, to induce non-linearity in the feature responses, and to incorporate luminance invariance (See later in Section \ref{sec:discussion}). Hence, the final set of {filter response} images can be denoted as 
\begin{equation} \label{eq:final_images}
Tanh(\alpha \tilde{I} _1), Tanh(\alpha \tilde{I} _2), ... ,Tanh(\alpha \tilde{I} _n).
\end{equation}
Note that the $\alpha$ parameter controls the degree of non-linearity in the output. The optimal values of parameters $\alpha$ and $n$ are found empirically in section \ref{sec:experiments}.
The final filter response images are now reduced in spatial dimension, through spatial max-pooling operations. The max-pooling is beneficial for two reasons: it reduces the dimensionality of the input; and it adds some translation invariance to the classifier. The class decision made by the CNN corresponds to the output node with the maximum value, as is usual. Note that, each CNN decision here is made on spike-events accumulated across $T$ seconds. since we have $kt_{vox}=T$. The CNN architecture used in all our experiments is shown in Table \ref{table:cnn_arch}. For all our experiments, we keep the CNN architecture unchanged. The CNN architecture used is kept quite shallow, as compared to state-of-the-art deep CNNs, due to the initial filtering stage already adding a layer of computation. }

 
\section{Experiments and Results}  \label{sec:experiments}


\begin{table}\label{table:cnn_arch}
\caption{CNN architecture used in our experiments. The number of filters in the first channel ($N$), and the number of categories in the last layer ($N_C$), are subject to change.}
\begin{tabular}{@{}c|ccccl@{}}
\toprule
Layer & \begin{tabular}[c]{@{}c@{}}Map Size \\ (After layer)\end{tabular} & Features                                                          & Layer Type                                                 & Kernel & Stride \\ \midrule
1            & 31x31                                                             & $N$                                            &                                                            &        &        \\
2            & 29x29                                                             & 60                                                                & conv                                                & 3x3    & 1      \\
3            & 15x15                                                             & 60                                                                & maxpool                                                   & 2x2    & 2      \\
4            & 13x13                                                             & 150                                                               & conv                                                & 3x3    & 1      \\
5            & 7x7                                                               & 150                                                               & maxpool                                                   & 2x2    & 2      \\
6            & 5x5                                                               & 300                                                               & conv                                                & 3x3    & 1      \\
7            & 3x3                                                               & 300                                                               & maxpool                                                   & 2x2    & 2      \\
8            & 1x1                                                               & 600                                                               & conv                                                & 3x3    & 1      \\
9            & 1x1                                                               & \begin{tabular}[c]{@{}c@{}}$N_{C}$\end{tabular} & \begin{tabular}[c]{@{}c@{}}Fully \\ Connected\end{tabular} & 1x1    & 1      \end{tabular}
\end{table}

The evaluation of our two-stage spatiotemporal filtering and classification framework is done through extensive testing on the IBM gesture recognition dataset \cite{gesture_ibm} and a novel action recognition dataset recorded for this work. However, our proposed framework contains free parameters that need to be fixed beforehand:

\begin{itemize}
\item {$N$ (Number of spatiotemporal filters):} Unless mentioned otherwise, $N=9$ for all our experiments. The only obvious exception being the analysis of performance w.r.t change of $N$ (Section \ref{sec:filter_number_gesture}).
\item {$\alpha$ (Sigmoid parameter):} This parameter was fixed through validation on the gesture recognition training data ($80/20$), for spike-event duration of 50 ms. From the validation results, we found that $\alpha = 0.2$ returned the best recognition accuracy. Hence, for all our experiments we keep $\alpha=0.2$.
\item {$a$ and $k$ (Spatiotemporal filter ROI size):} The values of the spatiotemporal ROI for the filters is decided subjectively, by observing the spatiotemporal filters returned by the algorithm. This is detailed in the next section. We find that choosing $a=6$, and $k=25$ returns spatiotemporal filters with most desirable properties. 
\item {$T$ (CNN decision duration): This represents the temporal duration of spike-event data chosen for a single CNN decision. In this work, we only experiment with $T=50$ ms and $T=100$ ms. When making class decisions on spike-event input across a time duration of $T_{total}>T$, we divide the total duration into overlapping segments, each of which has a duration of $T$. We aggregate each CNN decision on each segment, to obtain the final category estimate. This process will be explained in detail later in the appropriate sections.} 
\end{itemize}


%

To create the training examples, the same procedure mentioned in the beginning of section \ref{sec:methods} is followed. Each training example is generated from convolving $n$ filters of size $a \times a \times k$, on the voxellized event array created from events occuring within a time duration of $kt_{vox}$ seconds. For each analysis we describe the choice of parameters $a$, $k$ and $t_{vox}$. In both the datasets, the training examples were generated from event-data corresponding to a certain number of subjects (fixed). The other subjects were used for generating the testing data samples. 

We begin by analyzing the spatiotemporal filters that emerge from the optimization described in section \ref{sec:methods}. 

\begin{figure*}[htbp]
	\centering
	\includegraphics[width=0.9\textwidth]{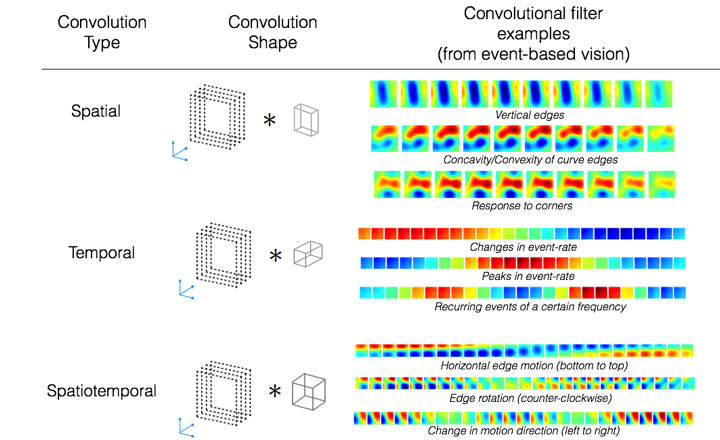}
	\caption{Exemplar spatiotemporal filters obtained from the proposed method. The filters are shown as a sequence of images, with each image revealing a flat cross-section of the 3D spatiotemporal filter. Successive images reflect the evolution of the filter weights in time. Three sets of selected filters are shown in this figure, corresponding to different shapes of the filter ROI (flat, long and moderate).}
	\label{fig:filter_shape_temporal}
\end{figure*}

\subsection{Spatiotemporal Filters: What is learned ?}

The spatiotemporal feature learning method proposed here will return spatiotemporal weights as convolutional filters of size $(a\times a\times k)$ which convolve on a spatiotemporal region spanning a volume of $(a$ pixels $\times a$ pixels $\times kt_{vox}$ ms $)$. Naturally, the filters returned by the optimization will show differences for different values of these parameters. To remind, our SFA like optimization objective enforces robustness to event removal related changes to the spike-event pattern. Here we wish to underline the various aspects of a visual spatiotemporal feature which are captured by the learned filters. Note that all these filters have been learned from the IBM gesture recognition dataset. 

{Following figure \ref{fig:filter_shape_temporal}, we document our observations as follows.} 

\begin{itemize}

\item When the filters are \textit{flatter} (k $\sim$ 10) and have a larger spatial size (a = 12), the learned weights primarily encode spatial structure. Examples are shown in Fig. \ref{fig:filter_shape_temporal}(a), where we find that certain filters respond the highest to edges which have a convex shape, and the most negative to concave ones. 
\item When the filters are \textit{longer} (k $\sim$ 40, Fig. \ref{fig:filter_shape_temporal} (b)) and have a smaller spatial size (a = 2), the learned weights primarily encode temporal structure. More precisely the weights now are analogous to temporal derivatives of various orders. 
\item When the filters are moderately long (k $\sim$ 25, Fig. \ref{fig:filter_shape_temporal} (c)) and have a moderate spatial size (a = 6), the learned weights encode both spatial and temporal structure. These filters are the ones with the most interesting properties, as some of them respond to edge motion, others only to spatial edges. More intriguingly, certain filters have maximum response to back-and-forth motion in horizontal and vertical directions. 
\end{itemize}


%
%

%

\subsection{Gesture Recognition: DVS Gesture Dataset}

The gesture recognition dataset recorded in \cite{gesture_ibm} contains a total of 10 categories of gestures, with one additional \textit{other gesture} class. For each subject, the gestures were recorded in a total of three lighting conditions. The training/testing split is fixed, such that the training contains 23 subjects, and the testing data is from the other 6. First, the spatiotemporal filters were learnt from the training data with the method described in Section \ref{sec:learn_filters}. Here we report the performance of the classifiers when all of the training data has been used to learn the spatiotemporal filters.


The previous section establishes that by only using appropriately sized spatiotemporal ROIs, one can learn filters which extract meaningful spatial (edges) and temporal (motion) information. All of our experiments henceforth are with a fixed spatiotemporal ROI size of $6 \times 6 \times 25$. 

{The following sections will evaluate the effects of CNN decision duration ($T$), number of filters ($N$), total input duration ($T_{total}$) and spike-time precision, on the test data recognition accuracy in the DVS Gesture dataset.} 

\subsubsection{$T = 50$ ms}
 

Here we fix the CNN decision duration to $T = 50$ ms. First, spatiotemporal weights are learned from the event-data recordings in the training data (23 subjects). Using validation tests, we decide the optimal value of $\alpha$, the coefficient which controls the non-linearity of the filter. We find that $\alpha =0.2$ results in best validation data accuracy, and therefore $\alpha$ is set to $0.2$ for all the remaining experiments. Here, the CNN achieves a 93.5\% recognition accuracy on the testing data. Note that this number is just 1\% shy of the state-of-the-art performance demonstrated in \cite{gesture_ibm}, which uses a total of 192 ms of spike-event time for each classification decision. 



%

\subsubsection{Number of filters ( $T=100$ ms )} \label{sec:filter_number_gesture}

\begin{figure}[htbp]
	\centering
	\includegraphics[width=0.45\textwidth]{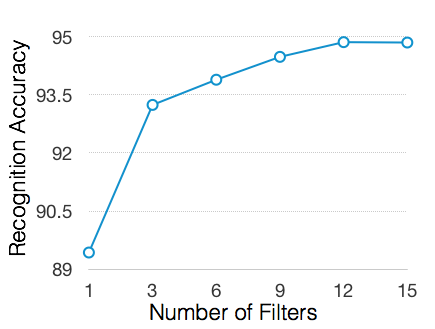}
	\caption{Recognition accuracy trend on the test examples of the gesture recognition dataset, as a function of the number of filters used in the initial pre-processing stage.  }
	\label{fig:filter_number}
\end{figure}

An important parameter in our approach is the number of spatiotemporal filters, N, which can be considered as being analogous to the number of hidden units in the second layer of the entire network. Choosing a high value of $N$ would ultimately result in a higher number of hidden parameters than the actual input, and therefore would be unadvisable. For this reason, we test the classifier performance when $N$ varies within the range of 1 to 15. Note that each time, only the $N$ best projections, according to the objective function in , are chosen as the spatiotemporal filters. {We choose a larger CNN decision duration ($T=100$ ms), and find that the CNN performance improves significantly than before, where $T=50$ ms.} The recognition accuracy obtained on the test subjects, as a function of $N$ is shown in Figure \ref{fig:filter_number}. The highest accuracy of 94.85\% is recorded for $N=?$, beyond which, the performance is found to reach a saturation point. In contrast, using a single spatiotemporal filter ($N=1$) only results in an 88\% recognition accuracy.

\subsubsection{With larger durations $(T_{total}>T)$}

To produce a fair comparison with state-of-the-art in this dataset, we need to obtain classification performance when 200 ms of spike-event data was used for each class decision. This is because the original accuracies reported in \cite{gesture_ibm} correspond to that case (T = 192 ms). Note that the accuracies we have obtained till now, correspond to the cases where 50 ms and 100 ms of input was used. To address this, here we document the classification performance of our method when different range of input spike-event durations were used for classification (till 500 ms). Each classification decision was obtained in a similar manner to the winner-take-all method used in \cite{gesture_ibm}. The "base" network used in our case was the one trained with 100 ms of input data and 12 input filter channels (previous section), which performed with a 94.85\% accuracy on the testing dataset. The final class decision over K ms (K$>$100) was obtained by placing consecutive overlapping windows of 100 ms (90 ms overlap), until the total duration of input covered by the windows equals K, and use the classification output of the CNN on each window. Subsequently, each window of spike-event data was channeled through the spatiotemporal filtering + CNN feedforward routine to obtain the output category for the corresponding window. The final output decision over K ms was considered to be the category which appeared the most in the system's output decisions for all the windows present. 

\begin{figure}[htbp]
	\centering
	\includegraphics[width=0.48\textwidth]{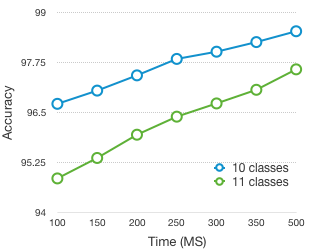}
	\caption{Figure showing classification performance with when different durations of input spike-event data was used for each classification decision. The two different plots correspond to the accuracies obtained for the cases when the "other gesture" class was removed (10 classes) and when all the classes were retained (11 classes). Naturally, performance improves with more input data, for both the cases. }
	\label{fig:gesture_acc_time}
\end{figure}

Figure \ref{fig:gesture_acc_time} shows the accuracy trends with varying input durations, for both the 10 class (without the "other gesture" class) and 11 class cases. For 200 ms wide input, we obtain a 11 class accuracy of 95.94\%, almost a 1.5\% improvement over the benchmark (94.5\%).{ It is clear that using larger $T_{total}$ duration for generating gesture class decisions leads to better performance. But however, note that a larger $T_{total}$ also increases the latency of the system in generating a class decision, making it considerably slower. The gesture cycles in the DVS Gesture dataset are quite short (200-250 ms), therefore for real-time recognition setting $T_{total}=250$ ms seems to be the prudent choice. }

\begin{figure}[htbp]
	\centering
	\includegraphics[width=0.48\textwidth]{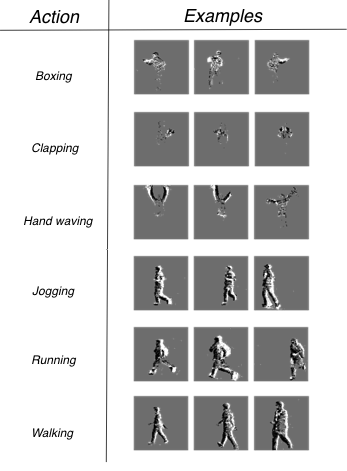}
	\caption{Figure showing selected binned spike-event data from all the categories of actions present in our recorded dataset. From these images, notice the variations of pose, distance and speed of movement. }
	\label{fig:action_dvs_images}
\end{figure}

{\subsubsection{Effect of changing spike-time precision}}

{In this analyses, we test the effect of spike-time precision on the overall accuracy and category-wise accuracy. From this evaluation, we intend to quantify the drop in accuracy, as spike-time precision is reduced. This will establish the degree of usefulness of the spike-timing precision in an event-based camera, with regards to gesture prediction with our two-stage architecture. In addition to overall accuracy changes, we also identify the categories which are affected the most by the the reduction of spike-time precision. }

{To artificially simulate reduction of timestamp precision, we add noise to each spike-event's timestamp, 
\begin{equation}
t_i  = t_i + \mathcal{U}(-t_{\delta}, t_{\delta}) 
\end{equation} 
}

{where the random variable $\mathcal{U}(a,b)$ represents a uniform distribution in the real line interval (a,b), and is responsible for adding noise and thereby reducing spike-time precision. We multiplicatively increase $t_{\delta}$, and re-run the trained networks on the noise-added spike-event data. The parameter values for these experiments are the same as previous, with the input duration being 100 ms, and the number of filters being 12 (as they produce the best performance). The results are summarized in Table \ref{table:st_precision}. Overall, we observe that the performance stays roughly stable for $t_{\delta}=2$ ms, but worsens significantly after that (4 ms onwards). A closer look at the individual category F1 scores reveals that only some of the categories (Drums, Guitar and Random Other Gestures) are affected in performance, while the others surprisingly maintain the accuracy level. We even observe that quite a few categories are unaffected by significant reduction ($t_{\delta} = 32$ ms) in temporal precision, demonstrating the fact that ultra-precise temporal information is not required for most gestures, except three in this case. The lasting lesson here is that fine spike-time precision does help in achieving better recognition  performance, only for the categories which contain intricate motion.}

\begin{table*}[]
\centering
\begin{tabular}{|l|l|l|l|l|l|l|}
\hline
\textbf{GESTURE CATEGORY} & \textbf{0 ms} & \textbf{2 ms} & \textbf{4 ms} & \textbf{8 ms} & \textbf{16 ms} & \textbf{32 ms} \\ \hline
\textbf{Hand Clapping} & 89.36 & 89.36 & 88.72 & 88.59 & 88.18 & 87.09 \\ \hline
\textbf{Right Hand Wave} & 99.10 & 99.10 & 98.91 & 98.62 & 98.66 & 98.33 \\ \hline
\textbf{Left Hand Wave} & 97.50 & 97.50 & 97.33 & 97.17 & 96.92 & 96.85 \\ \hline
\textbf{Right Hand Clockwise} & 97.05 & 97.05 & 97.03 & 96.85 & 96.65 & 95.99 \\ \hline
\textbf{Right Hand Counter Clockwise} & 97.90 & 97.90 & 97.72 & 97.50 & 97.29 & 96.79 \\ \hline
\textbf{Left Hand Clockwise} & 97.27 & 97.27 & 96.98 & 97.10 & 96.82 & 97.19 \\ \hline
\textbf{Left Hand Counter Clockwise} & 98.53 & 98.53 & 98.47 & 98.17 & 97.94 & 97.92 \\ \hline
\textbf{Forearm Roll (Forward/Backward)} & 98.48 & 98.48 & 98.25 & 98.22 & 98.01 & 97.80 \\ \hline
\textbf{Drums} & 87.04 & 87.04 & 85.85 & 85.73 & 85.43 & 84.50 \\ \hline
\textbf{Guitar} & 89.86 & 89.86 & 88.39 & 87.74 & 86.77 & 84.52 \\ \hline
\textbf{Other Random Gestures} & 81.64 & 81.64 & 80.39 & 79.84 & 78.00 & 73.96 \\ \hline
\textbf{OVERALL ACCURACY} & 94.80 & 94.78 & 94.31 & 94.10 & 93.70 & 92.90 \\ \hline
\end{tabular}
\caption{Table showing the F1 scores (harmonic mean of precision and recall) of each gesture category, and the overall recognition accuracy, on the 100 ms input duration IBM gesture recognition dataset. The columns indicate the extent of loss of spike-time precision}
\label{table:st_precision}
\end{table*}

\subsection{Action Recognition: New Dataset} 

A new dataset, comprising of six actions (see Fig. \ref{fig:action_dvs_images} , similar to KTH actions dataset for frame-based vision \cite{action_data_kth}), was recorded with the DVS sensor in fixed indoor lighting conditions. The participants were asked to incorporate more variation into their actions, such as speed, direction (for walking/jogging/running actions) and pose (for boxing/clapping/hand waving). The system was trained with the same parameters that resulted in the best recognition performance on the DVS Gesture dataset {($N=12$, $\alpha = 0.2$, $a=6, k=25$)}. Unlike the DVS Gesture dataset, the typical duration of an action cycle here can be much longer (1-2 seconds for walking). Thus, classifier decision based on 1-2 seconds worth of input will be used for validation, unlike in the previous case where only 100 ms of data was used to do the same. 

\subsubsection{With larger durations $(T_{total}>T)$} 


\begin{figure}[htbp]
	\centering
	\includegraphics[width=0.45\textwidth]{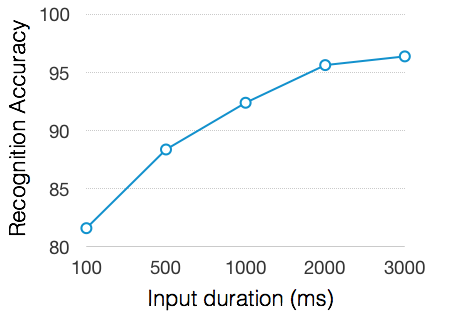}
	\caption{Recognition accuracy trend on the test examples of the action recognition dataset against change of input duration of spike-event data. }
	\label{fig:action_perf_duration}
\end{figure}

Here, we document the recognition performance, when spike-event data spread across various time durations is considered as the input to the system. Since each CNN decision is made with only $T=100$ ms of spike-event input, the final decision on the recognized action is made by aggregating multiple CNN output decisions made on overlapping segments of the spike-event data. The proportion of overlap is set at 50\%. The final class decision is decided by a winner-take-all mechanism on all CNN decisions made on each 100 ms segment. The results are shown in Figure \ref{fig:action_perf_duration}. We find that the recognition accuracy reaches 95.6\% for 2 seconds worth of spike-event input. {The gesture cycles in this dataset are longer (could reach 2 seconds), and therefore we choose $T_{total}=2$ seconds, for determining classifier accuracies on test data.} From the figure, we postulate that due to action cycles being longer, significant accuracy improvement is observed with larger durations.

\subsubsection{Comparing with PCA and Random Projections} \label{sec:pca_comparison}
 
 We compare performance of the two-stage architecture, when spatiotemporal filters other than our approach are used. Two other ways of generating spatiotemporal projections are experimented with, as follows:
\begin{itemize}
\item \textbf{PCA:} The top k principal components are used in the spatiotemporal filtering stage, instead of the SFA derived projections used by our method. 
\item \textbf{Random Projections:} We select k random projections, which are generated by the normal distribution. Each projection is normalized such that they become unit vectors.
\end{itemize}

\begin{table} 
\centering
\caption{Recognition accuracy results on the action dataset for different methods of computing spatiotemporal features}
\label{table:comparison}
\begin{tabular}{cc}
\hline
Method                                                                                    & \begin{tabular}[c]{@{}c@{}}Recognition \\ Accuracy\end{tabular} \\ \hline
\multicolumn{1}{c|}{PCA}                                                                  & 78.5\%                                                          \\
\multicolumn{1}{c|}{Random Projections}                                                   & 77.1\%                                                          \\
\multicolumn{1}{c|}{\begin{tabular}[c]{@{}c@{}}Our Approach\\ (Event-Removal + SFA)\end{tabular}} & 95.6\%                                                          \\ \hline
\end{tabular}
\end{table}

Results are tabulated in Table \ref{table:comparison}. For each method, the class decisions are generated on 2 seconds of data, using the winner-take-all mechanism described previously. As observed from the table, both PCA and random projections perform much worse than our method, almost showing a 20\% drop in recognition accuracy. 

\section{Discussions} \label{sec:discussion}

{A few significant examples of previous work in spatial or spatiotemporal filtering for event-based recognition problems include \cite{cnn_orientation_filters} and \cite{hfirst}. In both however, mainly hand-crafted orientation detection based filters are proposed. \cite{hfirst} mimics the visual cortex hierarchy by successively increasing filter complexity over each subsequent layer, before finally using a template matching based method for category estimation. The focus is primarily on extraction of spatial information. Similarly, in \cite{cnn_orientation_filters} an initial orientation detection layer is proposed, based on Gabor filters. The method in \cite{cnn_orientation_filters} is a two-stage network like ours, but the first layer contains hand-crafted spatial orientation detection filters, in contrast to ours.} 
 
{We re-iterate that this work is the first demonstration of a spatiotemporal filtering layer which extracts local spatiotemporal information from raw spike-event data, while limiting temporal precision loss. Furthermore, this layer is trained in an unsupervised manner, to demonstrate robustness (invariance) to the stochastic nature of event-based information in general, making our technique more generally applicable to other problems. Lastly, flexibility over the choice of filter number $N$, response non-linearity $\alpha$, and spatiotemporal size parameters ($a\times a \times T$), allows our method to cater to variations in dataset size, type of scenes (illumination variation levels), and also the type of information that one needs to encode (Fig. \ref{fig:filter_shape_temporal}). The performance of the two stage-model over the two datasets used in our experiments demonstrates the effectiveness of this method for classifying spatiotemporal patterns. }

{The accuracy plot in Fig. \ref{fig:filter_number} demonstrates that for the traffic dataset, around 12 filters produce the best performance, after which the accuracy reaches a plateau. We did not experiment with very high values of $N$, the number of spatiotemporal filters, because of the same. Furthermore, using too many filters increases the redundancy of the output of this layer, which also leads to very high dimensional input to the subsequent CNN (too many channels). A key difference between the performance on the two datasets lies in the typical gesture/action cycle length. Whereas for the gesture recognition data the accuracy reaches decent levels with only 100 ms of data (94.8\%), for the action recognition data the system requires longer durations to reach a similar level of performance (2 seconds  - 95\%). The reason for this is the longer duration of each action in the second dataset (boxing, hand-waving and walking), as opposed to the first one where most actions finish within 200 ms. Furthermore, the results in Table \ref{table:st_precision} demonstrates the importance of spike-time precision for correct classification. This was found to be especially true for a small subset of classes, which contain finer motions, which are indeed captured by the DVS, and exploited by our system.}

In what follows, we underline the key differences between our approach (spike-event removal invariance) and a PCA based approach (outlined in section \ref{sec:pca_comparison}), for spatiotemporal filter learning. The optimization in equation \ref{eq:to_minimize1} and \ref{eq:to_minimize2}  will return projections which have large variance overall (like PCA), conditioned on small variation between the projection response of $w^T \phi(X)$ and $w^TX$. A large number of factors can account for the variation of the spike-event patterns represented in $X$. One such example is the number of spike-events present within $X$ itself, which varies considerably in the data. This naturally implies that doing a PCA on $X$ will lead to projections which encode overall spike-event count within a spatiotemporal ROI.  

{However, here we present an argument for why the optimization method used here will tend to avoid such projections which explicitly encode spike-event count. Variations on spike-event count can occur due to variation in illumination and edge-sharpness. Observe that $\phi(X)$ is a many-to-one mapping between the spike-event set in $X$ and $\phi(X)$. Therefore, the number of spike-events in $\phi(X)$ will either be equal to, or less than the number of spike-events in $X$. As a result, we can think of $\phi(X)$ to be visually similar to the spike-event pattern present in $X$, but with fewer spike-events. Naturally, the filters returned by spike-event removal invariance optimization will avoid encoding spike-event count, as it is a significant source of variation between $X$ and $\phi(X)$. This is the key difference between our method and PCA\footnote{This fact can also be verified through the spatiotemporal filters themselves. We find that no filters explicitly encode spike-event count.}. The hyperbolic tangential non-linearity even further adds to the luminance-invariant and edge-sharpness invariant nature of the filter response images. These factors prove instrumental for good performance on both the datasets. This is because, both datasets contain considerable variation in illumination and edge-sharpness.  Subsequently, the non-linear transformation on the projection output (equation \ref{eq:final_images}) allows our generated projections to be robust to changes in the spike-event pattern due to such factors of variation.}

Along with the factors underlined above, there lies a philosophical aspect to training a system containing a two-stage, unsupervised-cum-supervised skeleton. Over the last two decades, hand-crafted features for classification have been almost entirely replaced by end-to-end training of a convolutional neural network {\cite{lowe_recognition,yann_midlevel,alexnet_imagenet}}. How could using hand-crafted features be detrimental to performance? In part, this can be answered by the information bottleneck criterion ( \cite{tishby_bottleneck,bottleneck_blackbox}). In simple words, the information bottleneck analysis states that each layer of a deep network must ideally strive to reduce category-irrelevant information present in the previous layer (compression), while avoiding the erasure of category-relevant information. So it is possible that using hand-crafted features affects the information bottleneck worse than the first layer of a CNN, which has been trained end-to-end. In other words, using hand-crafted pre-processing layer can remove considerable category-relevant details in addition to removing category-irrelevant information. 

So why not face the same problem in this work, where we also deploy spatiotemporal filtering as a pre-processing feature extraction stage ? We hypothesize that the answer to this question is two-fold. First, event-based data is very different from frame-based intensity images. It is much sparser, contains more noise, and due to the high temporal granularity is a lot more high dimensional in nature {\cite{dvs_original,temporal_ryad} }. These properties essentially point to there being an abundance of category-irrelevant information in the raw spike-event space. Therefore, using a preprocessing stage would be more welcome, and will help out the subsequent CNN classifier. Second, as we can choose the number of spatiotemporal filters, we can control the information bottleneck in a manner such that the optimal number of filters channel the most possible category-relevant information, and the least possible category-irrelevant information.

\section{Concluding Remarks}
This work proposes a novel two-stage unsupervised-supervised system for the categorization of spatiotemporal actions, with event-based input from a silicon retina. The first stage learns spatiotemporal convolutional filters using an SFA-like optimization routine. The obtained spatiotemporal filters are targeted to minimize event-removal related changes to a local spatiotemporal spike-event pattern. The second stage takes the output of the spatiotemporal filters as an input example containing multiple feature channels, and proceeds to train a classifier for recognition of spatiotemporal activity. For testing the system, two datasets are considered: DVS Gesture and a new action recognition dataset recorded for this work. Results demonstrate the ability of the system to outperform the state-of-the-art in event-based gesture recognition, along with demonstrating superior performance to other alternative ways of obtaining the first stage filters. We comment on the reasons entailing the effective performance of such a two-stage architecture, in the context of the event-based cameras, and also normal frame-based methods. {In general, we find that the spatiotemporal filtering layer is quite effective at extracting useful spatiotemporal information, as evidenced by the results and analyses, both in form and function. Such desirable properties enable our approach to scale to a wide array of problems, such as moving object tracking, motion segmentation and spatiotemporal feature matching. }




\bibliographystyle{IEEEtran}
\bibliography{main}

\begin{thebibliography}{10}
\providecommand{\url}[1]{#1}
\csname url@samestyle\endcsname
\providecommand{\newblock}{\relax}
\providecommand{\bibinfo}[2]{#2}
\providecommand{\BIBentrySTDinterwordspacing}{\spaceskip=0pt\relax}
\providecommand{\BIBentryALTinterwordstretchfactor}{4}
\providecommand{\BIBentryALTinterwordspacing}{\spaceskip=\fontdimen2\font plus
\BIBentryALTinterwordstretchfactor\fontdimen3\font minus
  \fontdimen4\font\relax}
\providecommand{\BIBforeignlanguage}[2]{{%
\expandafter\ifx\csname l@#1\endcsname\relax
\typeout{** WARNING: IEEEtran.bst: No hyphenation pattern has been}%
\typeout{** loaded for the language `#1'. Using the pattern for}%
\typeout{** the default language instead.}%
\else
\language=\csname l@#1\endcsname
\fi
#2}}
\providecommand{\BIBdecl}{\relax}
\BIBdecl

\bibitem{dvs_original}
P.~Lichtsteiner, C.~Posch, and T.~Delbruck, ``A 128 times; 128 120 db 15 956;s
  latency asynchronous temporal contrast vision sensor,'' \emph{IEEE Journal of
  Solid-State Circuits}, vol.~43, no.~2, pp. 566--576, Feb 2008.

\bibitem{atis_original}
C.~Posch, D.~Matolin, and R.~Wohlgenannt, ``An asynchronous time-based image
  sensor,'' in \emph{2008 IEEE International Symposium on Circuits and
  Systems}, May 2008, pp. 2130--2133.

\bibitem{davis_original}
C.~Brandli, R.~Berner, M.~Yang, S.~C. Liu, and T.~Delbruck, ``A 240 x00d7; 180
  130 db 3 x00b5;s latency global shutter spatiotemporal vision sensor,''
  \emph{IEEE Journal of Solid-State Circuits}, vol.~49, no.~10, pp. 2333--2341,
  Oct 2014.

\bibitem{cnn_digit}
P.~Y. Simard, D.~Steinkraus, and J.~C. Platt, ``Best practices for
  convolutional neural networks applied to visual document analysis,'' in
  \emph{Proceedings of the Seventh International Conference on Document
  Analysis and Recognition - Volume 2}, ser. ICDAR '03.\hskip 1em plus 0.5em
  minus 0.4em\relax Washington, DC, USA: IEEE Computer Society, 2003, pp.
  958--.

\bibitem{cnn_layerwise}
M.~Liu, J.~Shi, Z.~Li, C.~Li, J.~Zhu, and S.~Liu, ``Towards better analysis of
  deep convolutional neural networks,'' \emph{CoRR}, 2016.

\bibitem{hierarchy_recognition}
\BIBentryALTinterwordspacing
Y.~Guo, Y.~Liu, A.~Oerlemans, S.~Lao, S.~Wu, and M.~S. Lew, ``Deep learning for
  visual understanding: A review,'' \emph{Neurocomputing}, vol. 187, pp. 27 --
  48, 2016, recent Developments on Deep Big Vision. [Online]. Available:
  \url{http://www.sciencedirect.com/science/article/pii/S0925231215017634}
\BIBentrySTDinterwordspacing

\bibitem{cnn3d_action}
\BIBentryALTinterwordspacing
S.~Ji, W.~Xu, M.~Yang, and K.~Yu, ``3d convolutional neural networks for human
  action recognition,'' \emph{IEEE Trans. Pattern Anal. Mach. Intell.},
  vol.~35, no.~1, pp. 221--231, Jan. 2013. [Online]. Available:
  \url{http://dx.doi.org/10.1109/TPAMI.2012.59}
\BIBentrySTDinterwordspacing

\bibitem{cnn_sports1m}
\BIBentryALTinterwordspacing
A.~Karpathy, G.~Toderici, S.~Shetty, T.~Leung, R.~Sukthankar, and L.~Fei-Fei,
  ``Large-scale video classification with convolutional neural networks,'' in
  \emph{Proceedings of the 2014 IEEE Conference on Computer Vision and Pattern
  Recognition}, ser. CVPR '14.\hskip 1em plus 0.5em minus 0.4em\relax
  Washington, DC, USA: IEEE Computer Society, 2014, pp. 1725--1732. [Online].
  Available: \url{https://doi.org/10.1109/CVPR.2014.223}
\BIBentrySTDinterwordspacing

\bibitem{cnn3d_gesture}
P.~Molchanov, S.~Gupta, K.~Kim, and J.~Kautz, ``Hand gesture recognition with
  3d convolutional neural networks,'' 06 2015, pp. 1--7.

\bibitem{cnn_transferable_bengio}
J.~Yosinski, J.~Clune, Y.~Bengio, and H.~Lipson, ``How transferable are
  features in deep neural networks?'' in \emph{Proceedings of the 27th
  International Conference on Neural Information Processing Systems - Volume
  2}, ser. NIPS'14.\hskip 1em plus 0.5em minus 0.4em\relax Cambridge, MA, USA:
  MIT Press, 2014, pp. 3320--3328.

\bibitem{cnn_ts_steering}
A.~I. Maqueda, A.~Loquercio, G.~Gallego, N.~N. Garc{\'{\i}}a, and
  D.~Scaramuzza, ``Event-based vision meets deep learning on steering
  prediction for self-driving cars,'' \emph{CoRR}, vol. abs/1804.01310, 2018.

\bibitem{hots_garrick}
X.~Lagorce, G.~Orchard, F.~Gallupi, B.~E. Shi, and R.~Benosman, ``Hots: A
  hierarchy of event-based time-surfaces for pattern recognition,'' \emph{IEEE
  Transactions on Pattern Analysis and Machine Intelligence}, vol.~PP, no.~99,
  pp. 1--1, 2016.

\bibitem{mapping_frame_event}
J.~Perez-Carrasco, B.~Zhao, C.~Serrano, B.~Acha, T.~Serrano-Gotarredona,
  S.~Chen, and B.~Linares-Barranco, ``Mapping from frame-driven to frame-free
  event-driven vision systems by low-rate rate coding and coincidence
  processing--application to feedforward convnets,'' \emph{Pattern Analysis and
  Machine Intelligence, IEEE Transactions on}, vol.~35, no.~11, pp. 2706--2719,
  Nov 2013.

\bibitem{cnn_rohan}
R.~Ghosh, A.~Mishra, G.~Orchard, and N.~Thakor, ``Real-time object recognition
  and orientation estimation using an event-based camera and cnn,'' in
  \emph{Biomedical Circuits and Systems Conference (BioCAS), 2014 IEEE}, Oct
  2014, pp. 544--547.

\bibitem{cnn_action_separate}
\BIBentryALTinterwordspacing
Z.~Qiu, T.~Yao, and T.~Mei, ``Learning spatio-temporal representation with
  pseudo-3d residual networks,'' \emph{CoRR}, vol. abs/1711.10305, 2017.
  [Online]. Available: \url{http://arxiv.org/abs/1711.10305}
\BIBentrySTDinterwordspacing

\bibitem{cnn_3d_simple}
D.~Tran, L.~Bourdev, R.~Fergus, L.~Torresani, and M.~Paluri, ``Learning
  spatiotemporal features with 3d convolutional networks,'' in
  \emph{Proceedings of the 2015 IEEE International Conference on Computer
  Vision (ICCV)}, ser. ICCV '15.\hskip 1em plus 0.5em minus 0.4em\relax
  Washington, DC, USA: IEEE Computer Society, 2015, pp. 4489--4497.

\bibitem{gesture_ibm}
A.~Amir, B.~Taba, D.~Berg, T.~Melano, J.~Mckinstry, C.~Di~Nolfo, T.~Nayak,
  A.~Andreopoulos, G.~Garreau, M.~Mendoza, J.~Kusnitz, M.~Debole, S.~Esser,
  T.~Delbruck, M.~Flickner, and D.~Modha, ``A low power, fully event-based
  gesture recognition system,'' pp. 7388--7397, 07 2017.

\bibitem{action_data_kth}
\BIBentryALTinterwordspacing
C.~Schuldt, I.~Laptev, and B.~Caputo, ``Recognizing human actions: A local svm
  approach,'' in \emph{Proceedings of the Pattern Recognition, 17th
  International Conference on (ICPR'04) Volume 3 - Volume 03}, ser. ICPR
  '04.\hskip 1em plus 0.5em minus 0.4em\relax Washington, DC, USA: IEEE
  Computer Society, 2004, pp. 32--36. [Online]. Available:
  \url{http://dx.doi.org/10.1109/ICPR.2004.747}
\BIBentrySTDinterwordspacing

\bibitem{flow_ryad}
R.~Benosman, C.~Clercq, X.~Lagorce, S.~H. Ieng, and C.~Bartolozzi,
  ``Event-based visual flow,'' \emph{IEEE Transactions on Neural Networks and
  Learning Systems}, vol.~25, no.~2, pp. 407--417, Feb 2014.

\bibitem{cnn_orientation_filters}
H.~Wang, J.~Xu, Z.~Gao, C.~Lu, S.~Yao, and J.~Ma, ``An event-based
  neurobiological recognition system with orientation detector for objects in
  multiple orientations,'' \emph{Frontiers in Neuroscience}, vol.~10, p. 498,
  2016.

\bibitem{hfirst}
G.~Orchard, C.~Meyer, R.~Etienne{-}Cummings, C.~Posch, N.~V. Thakor, and
  R.~Benosman, ``Hfirst: {A} temporal approach to object recognition,''
  \emph{CoRR}, 2015.

\bibitem{motion_feature_event}
X.~Clady, J.-M. Maro, S.~Barré, and R.~B. Benosman, ``A motion-based feature
  for event-based pattern recognition,'' \emph{Frontiers in Neuroscience},
  vol.~10, p. 594, 2017.

\bibitem{sfa_original}
L.~Wiskott and T.~J. Sejnowski, ``Slow feature analysis: Unsupervised learning
  of invariances,'' \emph{Neural Comput.}, vol.~14, no.~4, pp. 715--770, Apr.
  2002.

\bibitem{lowe_recognition}
D.~G. Lowe, ``Object recognition from local scale-invariant features,'' in
  \emph{Proceedings of the Seventh IEEE International Conference on Computer
  Vision}, vol.~2, Sep. 1999, pp. 1150--1157 vol.2.

\bibitem{yann_midlevel}
Y.~Boureau, F.~Bach, Y.~LeCun, and J.~Ponce, ``Learning mid-level features for
  recognition,'' in \emph{2010 IEEE Computer Society Conference on Computer
  Vision and Pattern Recognition}, June 2010, pp. 2559--2566.

\bibitem{alexnet_imagenet}
\BIBentryALTinterwordspacing
A.~Krizhevsky, I.~Sutskever, and G.~E. Hinton, ``Imagenet classification with
  deep convolutional neural networks,'' \emph{Commun. ACM}, vol.~60, no.~6, pp.
  84--90, May 2017. [Online]. Available:
  \url{http://doi.acm.org/10.1145/3065386}
\BIBentrySTDinterwordspacing

\bibitem{tishby_bottleneck}
\BIBentryALTinterwordspacing
N.~Tishby and N.~Zaslavsky, ``Deep learning and the information bottleneck
  principle,'' \emph{CoRR}, vol. abs/1503.02406, 2015. [Online]. Available:
  \url{http://arxiv.org/abs/1503.02406}
\BIBentrySTDinterwordspacing

\bibitem{bottleneck_blackbox}
\BIBentryALTinterwordspacing
R.~Shwartz{-}Ziv and N.~Tishby, ``Opening the black box of deep neural networks
  via information,'' \emph{CoRR}, vol. abs/1703.00810, 2017. [Online].
  Available: \url{http://arxiv.org/abs/1703.00810}
\BIBentrySTDinterwordspacing

\bibitem{temporal_ryad}
H.~Akolkar, C.~Meyer, Z.~Clady, O.~Marre, C.~Bartolozzi, S.~Panzeri, and
  R.~Benosman, ``What can neuromorphic event-driven precise timing add to
  spike-based pattern recognition?'' \emph{Neural Computation}, vol.~27, no.~3,
  pp. 561--593, March 2015.

\end{thebibliography}
%
%
%

\end{document}